%% file: main.tex
\documentclass[10pt,twocolumn,letterpaper]{article}

\usepackage{iccv}              % To produce the CAMERA-READY version
\usepackage{multirow}
\usepackage{colortbl}
\usepackage{amsthm}
\newtheorem*{remark}{Remark}
\usepackage{tabularx}
\definecolor{myblue}{RGB}{47,85,151}
\definecolor{mygreen}{RGB}{84,130,53}
\usepackage{makecell}
\makeatletter
\def\blfootnote{\xdef\@thefnmark{}\@footnotetext}
\makeatother
\usepackage[accsupp]{axessibility}


\definecolor{iccvblue}{rgb}{0.21,0.49,0.74}
\usepackage[pagebackref,breaklinks,colorlinks,allcolors=iccvblue]{hyperref}

\def\paperID{4006} % *** Enter the Paper ID here
\def\confName{ICCV}
\def\confYear{2025}

\title{PacGDC: Label-Efficient Generalizable Depth Completion\\with Projection Ambiguity and Consistency}

\author{
\\[-20pt]
Haotian Wang\textsuperscript{1,2},
Aoran Xiao\textsuperscript{2},
Xiaoqin Zhang\textsuperscript{3}, 
Meng Yang\textsuperscript{1$^\dagger$},
Shijian Lu\textsuperscript{2$^\dagger$} \\[1pt]
\\[-5pt]
    $^1$Xi'an Jiaotong University,
    $^2$Nanyang Technological University, 
    $^3$Zhejiang University of Technology\\[-10pt]
}

\begin{document}
\twocolumn[{
\maketitle
\begin{center}
    \captionsetup{type=figure}
    \vspace{-5mm}
    \includegraphics[width=\linewidth]{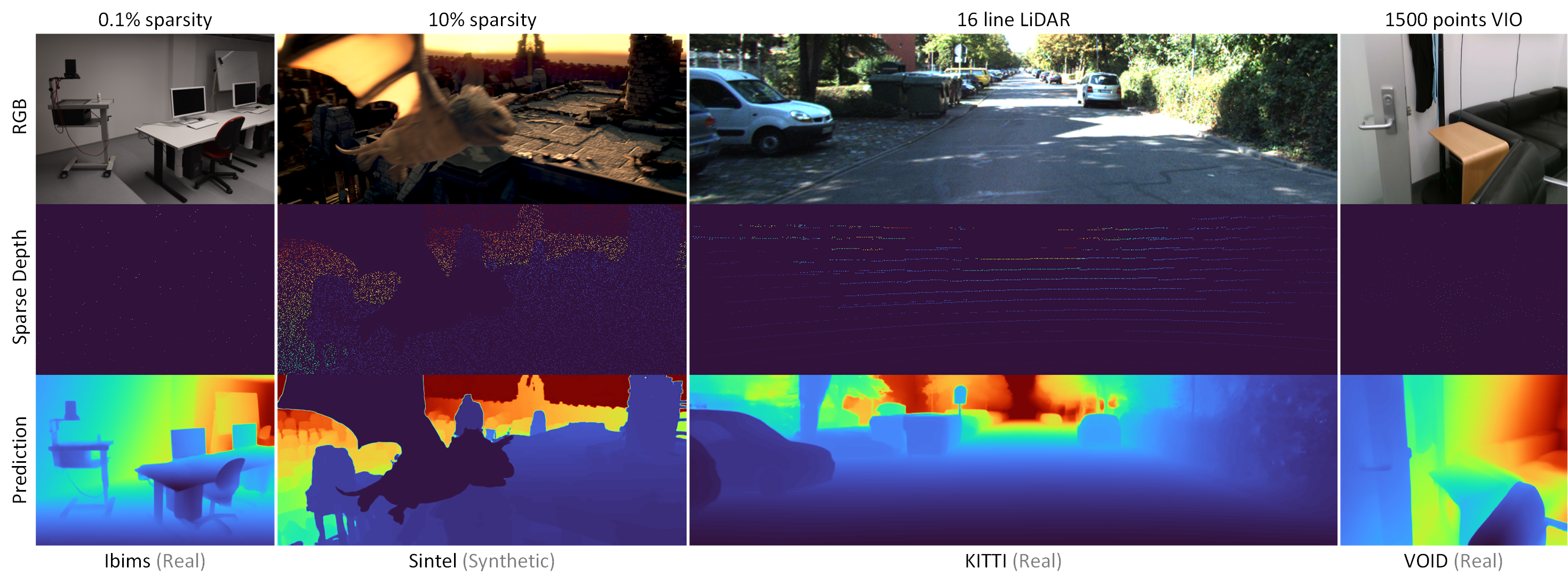}\\
    \captionof{figure}{PacGDC generalizes effectively across unseen scenarios with a wide range of \textit{scene semantics/scales} and \textit{depth sparsity/patterns}. The data include real ones from Ibims \cite{koch2018evaluation}, VOID \cite{wong2020unsupervised}, and KITTI \cite{geiger2012we}, as well as synthetic ones from Sintel \cite{butler2012naturalistic}. The sparse depths are captured from 0.1\%/10\% uniform sampling, 16 line vehicle LiDAR, and visual-inertial odometry (VIO) system with 1500 feature points.}
    \label{fig:teaser}
\end{center}
}]

\blfootnote{$^\dagger$Corresponding author.}

\input{sec/0_abstract}    
\input{sec/1_intro}
\input{sec/2_related}
\input{sec/3_method}
\input{sec/4_experiment}
\input{sec/5_conclusion}
\input{sec/6_acknowledgment}
{
    \small
    \bibliographystyle{ieeenat_fullname}
    \bibliography{main}
}

\input{sec/X_suppl}

\end{document}

%% file: sec/0_abstract.tex
\begin{abstract}
Generalizable depth completion enables the acquisition of dense metric depth maps for unseen environments, offering robust perception capabilities for various downstream tasks. However, training such models typically requires large-scale datasets with metric depth labels, which are often labor-intensive to collect.
This paper presents PacGDC, a label-efficient technique that enhances data diversity with minimal annotation effort for generalizable depth completion. PacGDC builds on novel insights into inherent ambiguities and consistencies in object shapes and positions during 2D-to-3D projection, allowing the synthesis of numerous pseudo geometries for the same visual scene. This process greatly broadens available geometries by manipulating scene scales of the corresponding depth maps. 
To leverage this property, we propose a new data synthesis pipeline that uses multiple depth foundation models as scale manipulators. These models robustly provide pseudo depth labels with varied scene scales, affecting both local objects and global layouts, while ensuring projection consistency that supports generalization. To further diversify geometries, we incorporate interpolation and relocation strategies, as well as unlabeled images, extending the data coverage beyond the individual use of foundation models.
Extensive experiments show that PacGDC achieves remarkable generalizability across multiple benchmarks, excelling in diverse scene semantics/scales and depth sparsity/patterns under both zero-shot and few-shot settings. Code: \url{https://github.com/Wang-xjtu/PacGDC}.
\end{abstract}

%% file: sec/1_intro.tex
\section{Introduction}
\label{sec:intro}

\begin{figure*}[!t]
  \centering
   \includegraphics[width=0.92\linewidth]{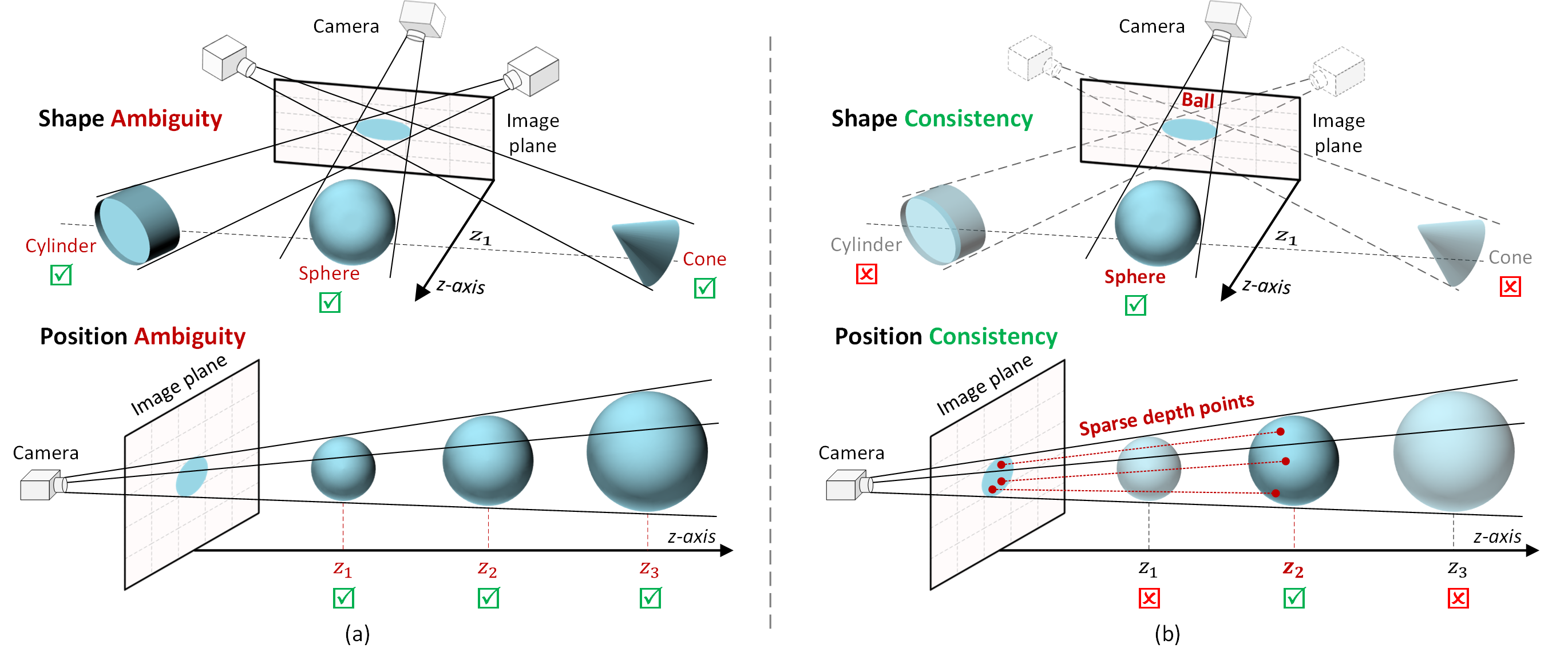}
   \caption{Illustration of the ambiguity and consistency in 2D-to-3D projections for generalizable depth completion. (a) \textit{Ambiguity}: shape ambiguity refers that the same 2D object can correspond to different 3D shapes, while position ambiguity refers that the same 3D shape can vary in size and position. (b) \textit{Consistency}: shape consistency denotes that the 3D shape aligns with the semantics of image input, while position consistency denotes that the 3D position is regularized by sparse depth input. The possible 3D objects are marked with \textcolor{green}{\checkmark}.
   }
   \label{fig:ambiguity&consistency}
\end{figure*}

Depth completion aims to infer dense metric depth maps from paired images and sparse depth measurements \cite{cheng2019learning}, providing accurate spatial representations that can support various downstream applications, such as robotics \cite{campos2021orb} and autonomous driving \cite{miao2023occdepth}. Despite its notable advancements, most existing methods suffer from poor generalization toward various new domains \cite{wang2023g2}. Recently, \textit{generalizable depth completion} resorts to domain-agnostic models that learn from source data but enable effective deployment across various unseen downstream environments \cite{wang2024scale, xu2024towards}. However, the success of these studies relies heavily on large-scale dense metric depth annotations to effectively cover real-world distributions, while collecting such annotated data is often laborious and time-consuming \cite{wang2023rgb}.

This paper introduces PacGDC, a label-efficient technique that is designed to maximize training data coverage with minimal annotation effort for generalizable depth completion. PacGDC is grounded in 2D-to-3D \textit{projection ambiguity}, where the same 2D image can be projected from multiple possible 3D geometric scenes \cite{yin2021virtual, ranftl2020towards}. 
To leverage this ambiguity, we decompose it into two orthogonal components: \textit{shape} and \textit{position}, as shown in \cref{fig:ambiguity&consistency}~(a). This decomposition reveals that each 3D geometry, defined by a depth map, can be uniquely identified by both shape and position cues. Meanwhile, these two cue types align well with the two input types in depth completion. As illustrated in \cref{fig:ambiguity&consistency}~(b), the shape cues (\textit{e.g.,} ``sphere") are consistent with semantic information (\textit{e.g.,} ``ball") in images, while sparse depth points help regularize spatial positions. Such \textit{consistency} mitigates ambiguity in generalizable depth completion, enabling accurate estimation of \textit{target metric depths} across unseen scenarios, as shown in \cref{fig:teaser}.
% We disentangle this into  \textit{shape ambiguity} and \textit{position ambiguity}, as shown in \cref{fig:ambiguity&consistency}~(a). This reveals that each 3D geometry, defined by a depth map, can be uniquely identified by both shape and position cues. Fortunately, object shape (\textit{e.g.} ``sphere") should be consistent with semantic information (\textit{e.g.} object ``ball") in images, while sparse depth points help regularize the spatial position, as illustrated in \cref{fig:ambiguity&consistency}~(b). Such \textit{consistency} alleviates the ambiguity in generalizable depth completion, enabling effective identification of \textit{target metric depths} across unseen scenarios, as shown in \cref{fig:teaser}.

\begin{figure*}[!t]
  \centering
   \includegraphics[width=\linewidth]{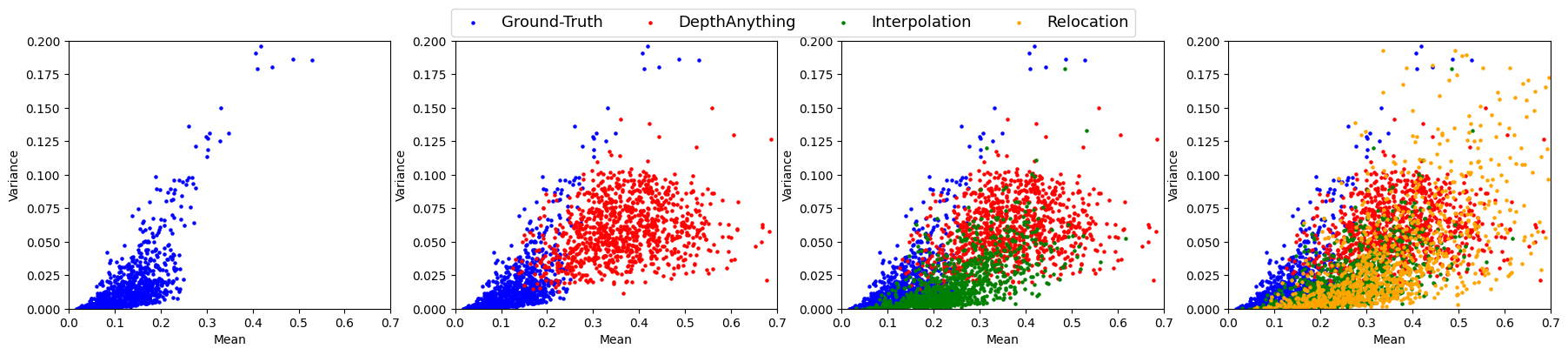}
   \caption{Data distribution of our synthesis method on 1000 samples from UnrealCV dataset \cite{wang2023g2}, visualized by mean and variance. From left to right, the data diversity increases progressively with each step. Notably, this statistic is based on a single foundation model, DepthAnything \cite{yang2024depth}, and a single instance of interpolation and relocation for simplicity. In the final implementation, multiple foundation models and randomized interpolation and relocation can be adopted to further enhance the data distribution.}
   \label{fig:diversity}
\end{figure*}

Building on these insights, this paper exploits these ambiguities to synthesize diverse pseudo geometries for the same visual scene, while maintaining consistencies among synthesized training triplets (\textit{i.e.,} images, sparse depths, and dense depth labels). It is achieved by manipulating scene scales of the corresponding depth maps, significantly expanding available geometries without requiring additional labeled samples. Since sparse depths with consistent positions can be sampled from dense labels, as introduced in \cite{wang2024scale, zuo2024omni}, our primary focus is to synthesize a large volume of pseudo dense depth labels that have consistent shapes. 

% Specifically, we exploit \textit{multiple foundation models of monocular depth estimation} to enhance the geometry diversity while maintaining shape consistency. 
Specifically, we exploit \textit{multiple foundation models of monocular depth estimation} to synthesize qualified depth labels. These models can robustly predict dense depth maps with consistent shapes/semantics from a single image \cite{yang2024depth, bochkovskii2024depth}, even across diverse unseen scenes. However, their predictions typically suffer from inaccurate scene scales, for both local objects and global layouts, due to the inherent scale ambiguity problem \cite{yin2021virtual, wofk2023monocular}. These characteristics allow generating pseudo dense depth maps that diverge from ground-truth labels in terms of scene scales, while preserving consistency in shape cues. To further diversify geometry, we incorporate interpolation and relocation strategies, enabling additional variations beyond the predictions of any individual foundation model. With the inclusion of unlabeled data, PacGDC significantly enriches the data diversity. The full synthesis pipeline is illustrated in \cref{fig:pipline}.

We conduct extensive experiments to validate the effectiveness of PacGDC in two practical applications: (1) \textit{zero-shot testing} on seven unseen datasets with diverse sparse depth inputs, including those captured from uniform sampling, visual-inertial odometry system, and vehicle LiDAR; (2) \textit{few-shot testing} on the KITTI dataset \cite{geiger2012we} with fewer than 1\% training data. The results show that our method achieves superior generalizable depth completion.

The major contributions of this paper are threefold:
\begin{itemize}
    \item We exploit a novel insight into 2D-to-3D projection ambiguity and consistency for generalizable depth completion, enriching data diversity without additional real labels.
    \item We propose a new data synthesis pipeline that manipulates local and global scene scales, enabling effective generalization to unseen domains with metric depths.
    \item Our approach achieves state-of-the-art performance in zero-/few-shot depth completion, validated across multiple benchmarks with different setups.
\end{itemize}

%% file: sec/2_related.tex
\section{Related Work}
\label{sec:relat}

\noindent\textbf{Depth Completion.} 
The success in deep learning has enabled depth completion methods to explore various dimensions including surface normal cues \cite{qiu2019deeplidar, xu2019depth, zhang2018deep}, semantics cues \cite{zhang2021multitask, nazir2022semattnet}, refinement strategies \cite{cheng2019learning, park2020non, wang2023lrru, tang2024bilateral}, advanced backbone architectures \cite{rho2022guideformer, shao2022towards, zhang2023completionformer}, and sophisticated feature fusion modules \cite{zhou2023bev, yan2024tri, long2024sparsedc, wang2024improving}. These approaches have effectively improved performance for intra-domain learning, where their training and testing data share similar scenes, such as KITTI \cite{geiger2012we} and NYUv2 \cite{silberman2012indoor}. However, they suffer degraded performance in unseen domains.

To address this limitation, G2-MonoDepth \cite{wang2023g2} explored a generalized framework for zero-shot scenarios. TDDC \cite{xu2024towards} incorporated pre-trained depth estimation models as a preprocessing step for enhanced image analysis. SPNet \cite{wang2024scale} investigated an important property of scale propagation within network architectures, while OMNI-DC \cite{zuo2024omni} introduced multi-resolution depth guidance.

In practical applications, a limited number of training samples can be collected for cost-efficient deployment in specific environments. Conventional intra-domain methods can be directly applied in such scenarios. Recently, UniDC \cite{park2024simple} explored few-shot depth completion using hyperbolic representation \cite{park2023learning}, while DDPMDC \cite{ran2023few} leveraged pre-trained diffusion models to mitigate data overfitting.

\noindent\textbf{Monocular Depth Estimation.} Existing monocular depth estimation methods can be broadly categorized into three main types. First, most methods focused on predicting metric depth maps within familiar training domains \cite{fu2018deep, yuan2022neural}, while they struggle to generalize to unseen domains. To improve robustness, some approaches predicted relative depth maps with normalized scales \cite{xian2020structure, ranftl2020towards, ranftl2021vision, ke2024repurposing, yang2024depth}, which disregarded scene scales for superior generalization. Recently, the community explored predicting metric depth maps from unseen scenes by incorporating the camera focal length \cite{yin2023metric3d, piccinelli2024unidepth, bochkovskii2024depth}. Our method manipulates the scene scales of depth maps using their dense predictions.

\noindent\textbf{Pseudo Labeling.} As a popular topic in semi-supervised learning, pseudo-labeling methods focused on leveraging unlabeled data by synthesizing pseudo labels \cite{li2023semireward}. The dominant methods aimed to improve pseudo label quality for reliable supervision, employing techniques such as threshold-based selection \cite{sohn2020fixmatch, wang2022freematch, zhang2021flexmatch, kim2022conmatch}, teacher-student frameworks \cite{zhou2010semi, xie2020self}, and advanced regularization strategies \cite{xie2020unsupervised, sohn2020fixmatch, li2021comatch}. These approaches have successfully enriched the training data of many foundational models \cite{kirillov2023segment, yang2024depth, ravi2024sam}. In contrast, our method investigates pseudo labels upon projection ambiguity for both labeled and unlabeled data.

%% file: sec/3_method.tex
\section{Method}
\label{sec:method}

\subsection{Problem Definition}
\label{subsec:definition}

\begin{figure*}[!t]
  \centering
   \includegraphics[width=\linewidth]{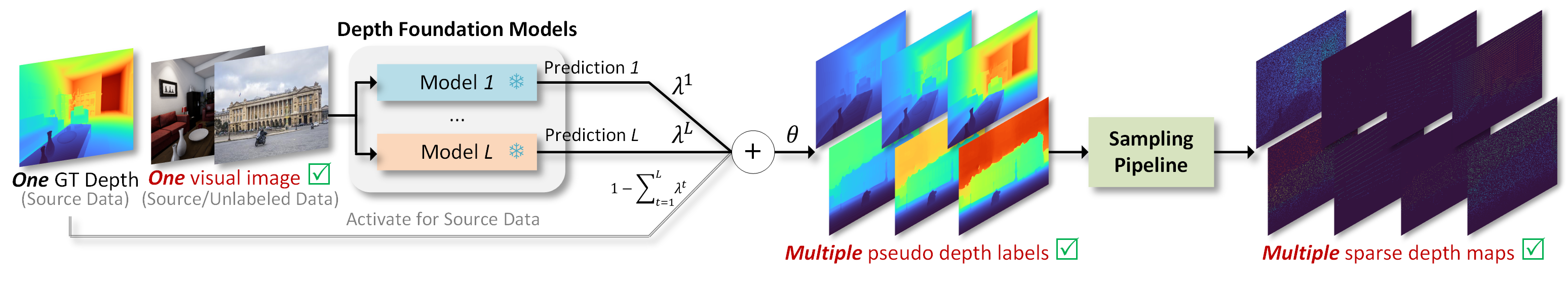}
   \caption{Overview of the proposed data synthesis pipeline, which leverages multiple depth foundation models, interpolation and relocation strategies, and unlabeled data. Sparse depth maps are then sub-sampled to form pseudo triplets (marked by \textcolor{green}{\checkmark}). This process significantly enhances data diversity through projection ambiguity while ensuring projection consistency that contributes to generalization.}
   \label{fig:pipline}
\end{figure*}

The training data of depth completion comprises annotated triplets, denoted $\mathcal{T} = \{I, p, d\}$, where $I$ represents the input image, $p$ is the sparse depth map captured by the depth sensors, and $d$ is the dense depth map of ground truth. The objective is to train a model $\mathcal{F}$, that can predict the dense depth map $\mathcal{F}(I, p)$ using both the input image and sparse depth map. The model is optimized by minimizing the difference between the predicted dense depth $\mathcal{F}(I, p)$ and the ground truth $d$, formulated as: $\min_{\mathcal{F}}|\mathcal{F}(I, p) - d|$.

In the context of \textit{generalizable} depth completion, the objective is updated to train the model $\mathcal{F}$ on source datasets with triplets $\mathcal{T}$, while enabling it to effectively generalize to unseen target data, achieving strong zero-shot performance.

This paper aims to achieve superior generalizable depth completion with minimal annotation effort. We develop a label-efficient solution, PacGDC, that synthesizes pseudo triplets $\hat{\mathcal{T}}$ to substitute for original triplets $\mathcal{T}$. This significantly enhances the diversity of source data, enabling better coverage for real-world data. 

Our approach addresses two main challenges:
\begin{itemize}
    \item The theoretical process of enhancing data diversity, which we tackle by leveraging the 2D-to-3D \textit{projection ambiguity and consistency}, as detailed in \cref{subsec:ambiguity}.
    \item The practical solution of synthesizing qualified data that contributes to generalization, for which we employ \textit{multiple depth foundation models}, as detailed in \cref{subsec:con&div}.
\end{itemize}

\subsection{Geometry Diversity from Projection Ambiguity and Consistency}
\label{subsec:ambiguity}

\textbf{Projection Ambiguity.} In the pin-hole camera model \cite{ wang2024dust3r}, the 2D coordinate $(u_i, v_i)$ in the image plane is mapped to their corresponding 3D position $(x_i, y_i, z_i)$ at pixel $i$ using the following projection relationship:
\begin{equation}
\label{eq:2dto3d}
    d_i P^{-1}
    \begin{bmatrix}
        u_i \\
        v_i \\
        1 \\
    \end{bmatrix}
    =
    \begin{bmatrix}
        x_i \\
        y_i \\
        z_i \\
    \end{bmatrix},
\end{equation}
where $P$ is the projection matrix and $d_i$ is the depth value at pixel $i$. By applying a scaling factor $\alpha_i$ to both sides of the equation, the same 2D coordinate $(u_i,v_i)$ can correspond to a new depth value $\hat{d}_i = \alpha_i d_i$ and a scaled 3D position $(\alpha_i x_i, \alpha_i y_i, \alpha_i z_i)$. This reveals that multiple 3D geometries can project onto the same 2D visual appearance $I$, by manipulating the scene scale of depth pixel $d_i$. This phenomenon is commonly referred to as \textit{projection ambiguity} or \textit{scale ambiguity} \cite{yin2021virtual, ranftl2020towards}.

In this paper, we decompose the projection ambiguity into two orthogonal sources, as illustrated in \cref{fig:ambiguity&consistency}~(a):
\begin{itemize}
    \item \textit{Shape ambiguity} refers to that one 2D object can correspond to different 3D shapes in the same position. This implies that corresponding depth maps may share similar means but differ in variances.
    \item \textit{Position ambiguity} means that the same 3D shape can vary in size and position. This suggests that their depth maps may have similar variances but varied means.
\end{itemize}
\begin{remark}
    These ambiguities indicate that each 3D geometry can be uniquely identified by both shape and position.
\end{remark}

\noindent\textbf{Projection Consistency.} The ambiguities suggest that predicting target 3D geometry, defined by the target depth map, requires identifying both its shape and position. In the training triplets $\mathcal{T}$, the input image $I$ provides semantic cues to identify the target shape. As illustrated in \cref{fig:ambiguity&consistency}~(b), the 2D object with semantic of \textit{``Ball"} should correspond to the 3D shape of \textit{``Sphere"}, rather than an unrealistic \textit{``Cone"} or \textit{``Cylinder"}. Meanwhile, the input sparse depth $p$ offers sparse depth points to regularize the target position.

These consistencies of shape and position are the foundation for achieving generalizable depth completion, where \textit{metric depths} can be effectively identified by these shape and position cues, even across diverse unseen scenarios. 

\noindent\textbf{Geometry Diversity}. Building on these insights, we leverage projection ambiguity to synthesize numerous pseudo geometries for the same visual scene, while maintaining projection consistency between the input data $I,p$ and the synthesized depth labels $\hat{d}$. This process can significantly enhance the geometry diversity of training data, thus achieving superior generalizability for depth completion.

We introduce the theoretical process of forming pseudo triplets $\hat{\mathcal{T}}$ to substitute for original triplets $\mathcal{T}$. First, we synthesize a set of dense depth maps $\{\hat{d}^j\}_{j=1}^{N}$, where $N$ is the number of pseudo depth labels per image. The specific synthesis method ensuring shape consistency is detailed in \cref{subsec:con&div}. Next, sparse depth maps are directly sub-sampled from each pseudo depth label, \textit{i.e.,} $\{\hat{p}^{j,k}\}_{j=1,k=1}^{N,M}$, where $M$ is the number of sparse depth maps per dense depth label. In this paper, we adopt the uniform sampling pipeline in \cite{wang2024scale} combined with the LiDAR\&SFM patterns sampling in \cite{zuo2024omni}. The sub-sampling naturally maintains consistent positions. Finally, the pseudo triplets consist of the original visual image, the pseudo depth labels, and the sampled sparse depth maps, \textit{i.e.,} $\hat{\mathcal{T}}=\{I,\{\hat{d}^j\}_{j=1}^{N},\{\hat{p}^{j,k}\}_{j=1,k=1}^{N,M}\}$. This process generates additional data combinations from the same visual images, beyond the original triplets $\mathcal{T}$.

\subsection{Qualified Synthesis with Scale Manipulation}
\label{subsec:con&div}

Our synthesis method aims to achieve two key objectives: (1) ensuring that the shape of the pseudo dense labels is consistent with the semantics of the input images, and (2) capturing diverse shapes and positions in the synthesized geometries. We leverage \textit{foundation models of monocular depth estimation} to accomplish the two goals.

\noindent\textbf{Pseudo Labels from Depth Foundation Model.} Depth foundation models, such as DepthAnything \cite{yang2024depth} and DepthPro \cite{bochkovskii2024depth}, are capable of robustly predicting dense depth maps with consistent semantic information from a single image, even across diverse unseen visual scenes. However, their predictions typically suffer from inaccurate scene scales due to the scale ambiguity inherent in single images \cite{ranftl2020towards}. For instance, the top-ranking depth estimation method \cite{piccinelli2024unidepth} on the KITTI leaderboard achieves 8.24 iRMSE, significantly weaker than the top-ranking depth completion method \cite{wang2024improving} with 1.83 iRMSE. Furthermore, although many methods adopt the global affine-invariant hypothesis, their predictions exhibit scale variations not only in the global layouts but also for local objects, as discussed in \cite{yin2021virtual, wofk2023monocular}. These characteristics enable the manipulation of local and global scene scales of pseudo depth labels to generate diverse geometries, while maintaining consistency in shape cues.

\begin{figure}[!t]
  \centering
   \includegraphics[width=\linewidth]{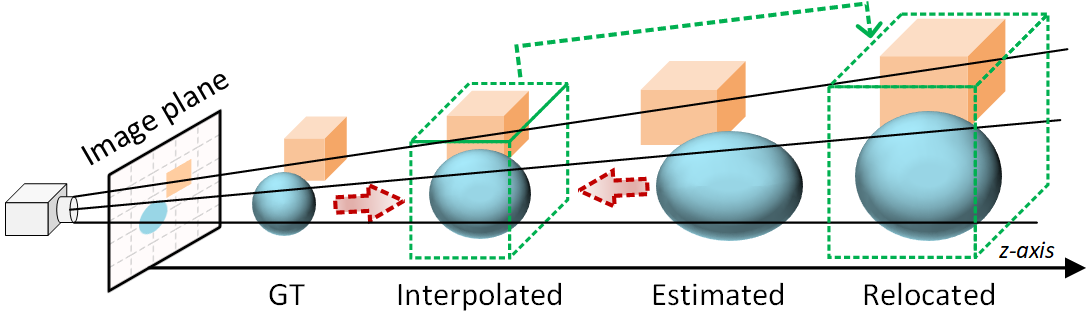}
   \caption{Illustration of basic geometry synthesis.}
   \label{fig:geometry_syn}
\end{figure}

Therefore, we denote the pseudo depth labels generated by the depth foundation model $\mathcal{R}$ from visual inputs $I$ as $\hat{d}=\mathcal{R}(I)$. Unfortunately, a single $\mathcal{R}$ only generates one type of dense prediction $\mathcal{R}(I)$, depending on its network architecture, training data, and training strategy. To diversify synthesized geometries, we further incorporate interpolation and relocation strategies. First, we randomly interpolate the original ground-truth depth maps $d$ with pseudo dense labels $\mathcal{R}(I)$. This operation can fill the geometry coverage between them, as illustrated in \cref{fig:geometry_syn}. One limitation is that the filled coverage depends on the initial spatial positions of the two dense maps. To address this, we randomly relocate the interpolated results into new positions. The pseudo depth labels are formalized as:
\begin{equation}
\label{eq:inter&reloc}
\hat{d} = \theta (\lambda \mathcal{R}(I) + (1 - \lambda) d),
\end{equation}
where $\lambda$ and $\theta$ are the random factors for interpolation and relocation, respectively. The diversity introduced by these strategies is demonstrated in \cref{fig:diversity}, and their effects are also verified in the ablation study in \cref{subsec:ab}.

\noindent\textbf{Extensions to Multiple Foundation Models and Unlabeled Images.} The basic synthesis pipeline can be extended by using multiple depth foundation models and incorporating unlabeled images. First, instead of relying on a single model $\mathcal{R}$, we update it to a set of models $\{\mathcal{R}^{t}\}_{t=1}^{L}$, where $L$ denotes the number of depth foundation models. This expansion increases the variety of pseudo dense labels, yielding $\mathcal{R}^{t}(I)$. By combining these models with the interpolation and relocation strategies from \cref{eq:inter&reloc}, the pseudo depth labels are updated as follows:
\begin{equation}
\label{eq:mulit-models}
\hat{d} = \theta \left({\sum_{t=1}^L}\lambda^{t} \mathcal{R}^{t}(I) + (1 - {\sum_{t=1}^L}\lambda^{t}) d\right),
\end{equation}
where $\lambda^{t}$ is random interpolation factor for each foundation model, with the constraint ${\sum_{t=1}^L}\lambda^{t} \leq 1$. This paper adopts two foundation models with quite different designs as examples, including DepthAnything \cite{yang2024depth} and DepthPro \cite{bochkovskii2024depth}.

As discussed in \cref{subsec:ambiguity}, PacGDC emphasizes assigning multiple pseudo depth labels to a single visual scene. This design shifts the model’s focus from regular \textit{fitting dataset priors} to ours \textit{learning geometric alignment}. It suggests that pseudo data, even without ground truth scene scale, can still contribute effectively to training generalizable depth completion models.  This insight motivates us to incorporate unlabeled images $I^u$ into our synthesis pipeline, further enriching data diversity from additional semantics and scene scales. Consequently, the set of visual images expands to $\hat{I}=\{I,I^u\}$. In this study, we incorporate 390K unlabeled images from SA1B dataset \cite{kirillov2023segment} for validation.

% Based on the analysis of projection ambiguity and consistency in \cref{subsec:ambiguity}, we argue that the ground-truth scene scale may not be crucial for training generalizable depth completion models. This insight allows us to incorporate unlabeled images $I^u$ into our synthesis pipeline, further enriching geometry diversity from additional semantics and scene scales. Consequently, the set of visual images expands to $\hat{I}=\{I,I^u\}$. In this study, we incorporate 390K unlabeled images from SA1B dataset \cite{kirillov2023segment} for validation.

\begin{remark}
    The final triplets $\hat{\mathcal{T}}$ are constructed by three elements: visual images $\hat{I}$, pseudo dense labels $\hat{d}$ generated by \cref{eq:mulit-models} from $\hat{I}$
    \footnote{\cref{eq:mulit-models} is available for unlabeled images by setting ${\sum_{t=1}^L}\lambda^{t}=1$}
    , and sparse depth maps $\hat{p}$ sampled from $\hat{d}$.
\end{remark}
The synthesis pipeline is illustrated in \cref{fig:pipline}, which significantly enhances data diversity without requiring any additional real annotations. The ablation study of these strategies is provided in \cref{subsec:ab}.

\begin{table}[!t]
\centering
\scriptsize
\resizebox{\linewidth}{!}{
\begin{tabular}{lc@{\hskip 5pt}ccc}
\toprule
Datasets & Indoor & Outdoor & Label & Size \\
\midrule
Matterport3D \cite{chang2017matterport3d} & \checkmark & & RGB-D & 194K \\
HRWSI \cite{xian2020structure} & \checkmark & \checkmark & Stereo & 20K \\
VKITTI \cite{gaidon2016virtual} & & \checkmark & Synthetic & 21K \\
UnrealCV \cite{wang2023g2} & \checkmark & \checkmark & Synthetic & 5K \\
BlendedMVS \cite{yao2020blendedmvs} & \checkmark & \checkmark & Stereo & 115K \\
SA1B \cite{kirillov2023segment} \textcolor{gray}{(subset)} & \checkmark & \checkmark & \textbf{None} & 390K \\
\bottomrule
\end{tabular}
}
\caption{The details of training datasets.}
\label{tab:trainingdataset}
\end{table}

\begin{table*}[!t]
\centering
\scriptsize
\resizebox{\linewidth}{!}{
\begin{tabular}{lc@{\hskip 5pt}cc@{\hskip 5pt}cc@{\hskip 5pt}cc@{\hskip 5pt}cc@{\hskip 5pt}cc@{\hskip 5pt}cc@{\hskip 5pt}c}
\toprule
 \multirow{2}{*}{Methods}  & \multicolumn{2}{c}{ETH3D} & \multicolumn{2}{c}{Ibims} & \multicolumn{2}{c}{NYUv2} & \multicolumn{2}{c}{DIODE} & \multicolumn{2}{c}{Sintel} & \multicolumn{2}{c}{KITTI} & \multicolumn{2}{c} {\textbf{Mean $\downarrow$ }} \\ 
\cmidrule(r){2-3} \cmidrule(r){4-5} \cmidrule(r){6-7} \cmidrule(r){8-9} \cmidrule(r){10-11} \cmidrule(r){12-13} \cmidrule(r){14-15}
  & \textit{RMSE} & \textit{MAE} & \textit{RMSE} & \textit{MAE} & \textit{RMSE} & \textit{MAE} & \textit{RMSE} & \textit{MAE} & \textit{RMSE} & \textit{MAE} & \textit{RMSE} & \textit{MAE} & \textit{RMSE} & \textit{MAE} \\ 
\midrule
LRRU \cite{wang2023lrru} & 5226 & 3932 & 796  & 599 & 1924 & 1387 & 9014 & 5714 & 18250 & 13249 & 10783 & 7450 & 7665 & 5389 \\
ImprovingDC \cite{wang2024improving} & 36479 & 5231 & 1021 & 845 & 2279 & 1809 & 9451 & 6562 & 18256 & 13787 & 11705 & 8396 & 8198 & 6105 \\
NLSPN \cite{park2020non}  & 2283 & 1367 & 239 & 116 & 414 & 210 & 5172 & 2379 & 43424 & 34221 & 4170 & 1911 & 9284 & 6701 \\
CFormer \cite{zhang2023completionformer}  & 1821 & 810 & 215 & 71 & 421 & 174 & 5176 & 2197 & 26415 & 21807 & 4400 & 1960 & 6408 & 4503 \\
G2MD \cite{wang2023g2}  & 1420 & 691 & 196 & 58 & 382 & 150 & 5026 & 2114 & 3612 & 917 & 3690 & 1607 & 2387 & 923 \\
SPNet \cite{wang2024scale}  & 1544 & 569 & 177 & 47 & 399 & 154 & 5070 & 2078 & \underline{3312} & \underline{659} & 3124 & 1240 & \underline{2271} & \underline{791} \\
OMNI-DC \cite{zuo2024omni}  & \underline{929} & \textbf{420} & \underline{165} & \underline{46} & \textbf{357} & \textbf{139} & \underline{4848} & \underline{2076} & 7733 & 3989 & \underline{3050} & \underline{1191} & 2847 & 1310\\
\rowcolor{gray!15} \textbf{Ours}  & \textbf{907} & \underline{454} & \textbf{160} & \textbf{46} & \underline{376} & \underline{147} & \textbf{4721} & \textbf{1984} & \textbf{2961} & \textbf{580} & \textbf{2673} & \textbf{1172} & \textbf{1966} & \textbf{731} \\
\bottomrule
\end{tabular}
}
\caption{\textbf{Zero-shot depth completion} on the six datasets with sparse depth maps obtained by uniformly sampling 10\%/1\%/0.1\% valid pixels. The \textbf{bold} indicates the best result, and the \underline{underline} indicates the second-best result.}
\label{tab:zs-unf}
\end{table*}

\begin{table*}[!t]
\centering
\scriptsize
\resizebox{\linewidth}{!}{
\begin{tabular}{lc@{\hskip 5pt}cc@{\hskip 5pt}cc@{\hskip 5pt}cc@{\hskip 5pt}cc@{\hskip 5pt}cc@{\hskip 5pt}cc@{\hskip 5pt}c}
\toprule
\multirow{2}{*}{Methods}  & \multicolumn{2}{c}{VOID-1500} & \multicolumn{2}{c}{VOID-500} & \multicolumn{2}{c}{VOID-150} & \multicolumn{2}{c}{KITTI-64L} & \multicolumn{2}{c}{KITTI-16L} & \multicolumn{2}{c}{KITTI-4L} & \multicolumn{2}{c}{\textbf{Mean $\downarrow$}}   \\
\cmidrule(r){2-3} \cmidrule(r){4-5} \cmidrule(r){6-7} \cmidrule(r){8-9} \cmidrule(r){10-11} \cmidrule(r){12-13} \cmidrule(r){14-15}
  & \textit{RMSE} & \textit{MAE} & \textit{RMSE} & \textit{MAE} & \textit{RMSE} & \textit{MAE} & \textit{RMSE} & \textit{MAE} & \textit{RMSE} & \textit{MAE} & \textit{RMSE} & \textit{MAE} & \textit{RMSE} & \textit{MAE} \\ 
\midrule
LRRU \cite{wang2023lrru}                                        & 835         & 530        & 937         & 637       & 1000        & 685       & 2124       & 742        & 4968       & 2943       & 12658      & 9466      & 3753                  & 2501                 \\
ImprovingDC \cite{wang2024improving}                                        & 950         & 595        & 1160        & 850       & 1251        & 949       & 2200       & 1118       & 4281       & 2325       & 11421      & 8044      & 3544                  & 2313                 \\
NLSPN \cite{park2020non}                                        & 431         & 156        & 484         & 192       & 571         & 247       & 1627       & 501        & 2174       & 711        & 4133       & 1690      & 1570                  & 583                  \\
CFormer \cite{zhang2023completionformer}                                     & 426         & 144        & 460         & 170       & 522         & 208       & 1513       & 359        & 2221       & 601        & 4768       & 1835      & 1652                  & 553                  \\
G2MD \cite{wang2023g2}                                        & 383         & 117        & 417         & 141       & 484         & 181       & 1570       & 352        & 2138       & 572        & 3941       & 1648      & 1489                  & 502                  \\
SPNet \cite{wang2024scale}                                       & \underline{353}         & \underline{104}        & \underline{375}         & \underline{119}       & \underline{430}         & \underline{151}       & 1523       & \underline{331}        & 2108       & 537        & 3268       & 1148      & 1343                  & 398                  \\
OMNI-DC \cite{zuo2024omni}                                     & 391         & 121        & 422         & 143       & 478         & 177       & \textbf{1191}       & \textbf{270}        & \textbf{1682}       & \textbf{441}        & \underline{2981}       & \underline{997}       & \underline{1191}                  & \underline{358}                \\
\rowcolor{gray!15} \textbf{Ours}                                     & \textbf{348}         & \textbf{102}        & \textbf{363}         & \textbf{114}       & \textbf{409}         & \textbf{141}       & \underline{1375}       & 337        & \underline{1702}       & \underline{460}        & \textbf{2685}       & \textbf{896}       & \textbf{1147}                  & \textbf{342}                \\
\bottomrule
\end{tabular}
}
\caption{\textbf{Zero-shot depth completion} on VOID dataset with 1500/500/150 sparsity levels from visual-inertial odometry system, and KITTI dataset with 64/16/4 beam lines from vehicle LiDAR.}
\label{tab:zs-lidar}
\end{table*}

\subsection{Learning from Synthesized Triplets}
\label{subsec:supervision}

Learning from such large-scale, diverse, and ambiguous training data presents a challenge for existing depth completion frameworks. To effectively leverage our synthesized data, we integrate our data settings into the SPNet \cite{wang2024scale} framework, known for its efficiency and strong generalization ability. The source training datasets, detailed in \cref{tab:trainingdataset}, include 355K labeled samples and 390K unlabeled samples.

\noindent\textbf{Computational Cost Analysis.} Since our work focuses on training data diversity, it does not introduce any additional computational cost during inference. This ensures that our model fully retains the efficient inference of SPNet, whose ``Tiny" model achieves 126.6 image/s on a single 3090 GPU at 320$\times$320 resolution. In comparison, competing methods such as G2-MonoDepth \cite{wang2023g2} and OMNI-DC \cite{bochkovskii2024depth} achieve only 69.5\% and 8.4\% of SPNet’s speed, respectively. 

The computational cost is primarily introduced during training. Our method consists of four main components: depth foundation models, unlabeled images, interpolation, and relocation. The latter two introduce only minimal additional multiplication and addition operations. For depth foundation models, their predictions can be precomputed and loaded on demand. Therefore, the primary additional cost derives from unlabeled images, resulting in an additional 390K/355K computations in our implementation.

%% file: sec/4_experiment.tex
\section{Experiment}
\label{sec:exp}
Our experiments consider two practical applications: \textit{zero-shot depth completion} in \cref{subsec:zero-shot} and \textit{few-shot depth completion} in \cref{subsec:few-shot}. Additional experimental results are included in the supplementary materials.

\subsection{Settings}
\noindent\textbf{Evaluation Protocol.} \textit{Zero-shot depth completion}: Following zero-shot depth estimation \cite{yang2024depth}, we impose no restrictions on the model training, evaluating only released models on the same test setups. The test datasets include two types:  (1) sparse depth maps obtained by uniformly sampling 10\%/1\%/0.1\% of valid pixels from the ETH3D \cite{schops2017multi}, Ibims \cite{koch2018evaluation}, NYUv2 \cite{silberman2012indoor}, DIODE \cite{vasiljevic2019diode}, Sintel \cite{butler2012naturalistic}, and KITTI \cite{geiger2012we} datasets, as used in \cite{wang2023g2,wang2024scale}; (2) sparse depth points captured by visual-inertial odometry system with 1500/500/150 sparsity levels on VOID \cite{wong2020unsupervised} dataset, and by vehicle LiDAR with 64/16/4 beam lines on KITTI dataset. \textit{Few-shot depth completion}: All models are trained on sequentially selected subsets of 1, 10, 100, and 1000 samples from the KITTI training set (\textit{i.e.,} 86K samples), and evaluated on its validation set with 1000 test samples.

\noindent\textbf{Implementation Details.} \textit{Zero-shot depth completion}: We adopt the ``Large" model of SPNet for the best performance, while sacrificing 47\% inference speed. The model is trained using the AdamW optimizer with batch size 192, running on six 3090 GPUs. The initial learning rate is 0.0002 with cosine learning rate decay using 100 epochs. \textit{Few-shot depth completion}: Our models are initialized by pre-trained weights from the zero-shot phase. We adopt 1/10 initial learning rate for this fine-tuning. The batch sizes are set to \{1, 1, 4, 4\} when using \{1, 10, 100, 1000\} training samples.

\begin{table}[!t]
\centering
\resizebox{\linewidth}{!}{
\begin{tabular}{llcccc}
\toprule
Shot               & Methods  & \textit{RMSE} $\downarrow$ & \textit{MAE} $\downarrow$ & \textit{iRMSE} $\downarrow$ & \textit{iMAE} $\downarrow$ \\
\midrule
\multirow{4}{*}{\textit{1}}    & LRRU \cite{wang2023lrru}    & 2138 & 679 & 95.69 & 2.94 \\
                      & ImprovingDC  \cite{wang2024improving} & 1358 & 337 & 4.68  & 1.43 \\
                      & SparseDC \cite{long2024sparsedc} & 1757 & 636 & 8.30  & 3.65 \\
                      & \textcolor{gray}{UniDC \cite{park2024simple}} & \textcolor{gray}{1684} & \textcolor{gray}{522} & \textcolor{gray}{-}  & \textcolor{gray}{-} \\
                      & \cellcolor{gray!15} \textbf{Ours}   & \cellcolor{gray!15} \textbf{1078} & \cellcolor{gray!15} \textbf{250} & \cellcolor{gray!15} \textbf{2.90}  & \cellcolor{gray!15} \textbf{0.98} \\
\midrule
\multirow{4}{*}{\textit{10}}   & LRRU  \cite{wang2023lrru} & 1337 & 342 & 5.54  & 1.38 \\
                      & ImprovingDC   \cite{wang2024improving} & 1316 & 315 & 4.13  & 1.26 \\
                      & SparseDC \cite{long2024sparsedc} & 1438 & 380 & 10.22 & 1.86 \\
                      & \textcolor{gray}{UniDC \cite{park2024simple}} & \textcolor{gray}{1385} & \textcolor{gray}{407} & \textcolor{gray}{-}  & \textcolor{gray}{-} \\
                      & \cellcolor{gray!15} \textbf{Ours}  & \cellcolor{gray!15} \textbf{969} & \cellcolor{gray!15} \textbf{238} & \cellcolor{gray!15} \textbf{2.70}  & \cellcolor{gray!15} \textbf{0.96} \\
\midrule
\multirow{4}{*}{\textit{100}}  & LRRU \cite{wang2023lrru}   & 1261 & 295 & 4.02  & 1.17 \\
                      & ImprovingDC  \cite{wang2024improving}  & 1241 & 304 & 4.01  & 1.27 \\
                      & SparseDC \cite{long2024sparsedc} & 1203 & 325 & 4.42  & 1.48 \\
                      & \textcolor{gray}{UniDC \cite{park2024simple}} & \textcolor{gray}{1224} & \textcolor{gray}{339} & \textcolor{gray}{-}  & \textcolor{gray}{-} \\
                      & \cellcolor{gray!15} \textbf{Ours}   & \cellcolor{gray!15} \textbf{911}  & \cellcolor{gray!15} \textbf{229} & \cellcolor{gray!15} \textbf{2.54}  & \cellcolor{gray!15} \textbf{0.96} \\
\midrule
\multirow{4}{*}{\textit{1000}} & LRRU \cite{wang2023lrru}   & 1105 & 266 & 3.34  & 1.09 \\
                      & ImprovingDC  \cite{wang2024improving}   & 1121 & 279 & 3.79  & 1.15 \\
                      & SparseDC \cite{long2024sparsedc} & 1049 & 263 & 3.57  & 1.14 \\
                      & \cellcolor{gray!15} \textbf{Ours}  & \cellcolor{gray!15} \textbf{830}  & \cellcolor{gray!15} \textbf{220} & \cellcolor{gray!15} \textbf{2.28}  & \cellcolor{gray!15} \textbf{0.91} \\
\bottomrule
\end{tabular}
}
\caption{\textbf{Few-shot depth completion} on KITTI with 64 line LiDAR using 1, 10, 100, and 1000 training samples.}
\label{tab:fs-64line}
\end{table}

\begin{figure}[!t]
  \centering
   \includegraphics[width=\linewidth]{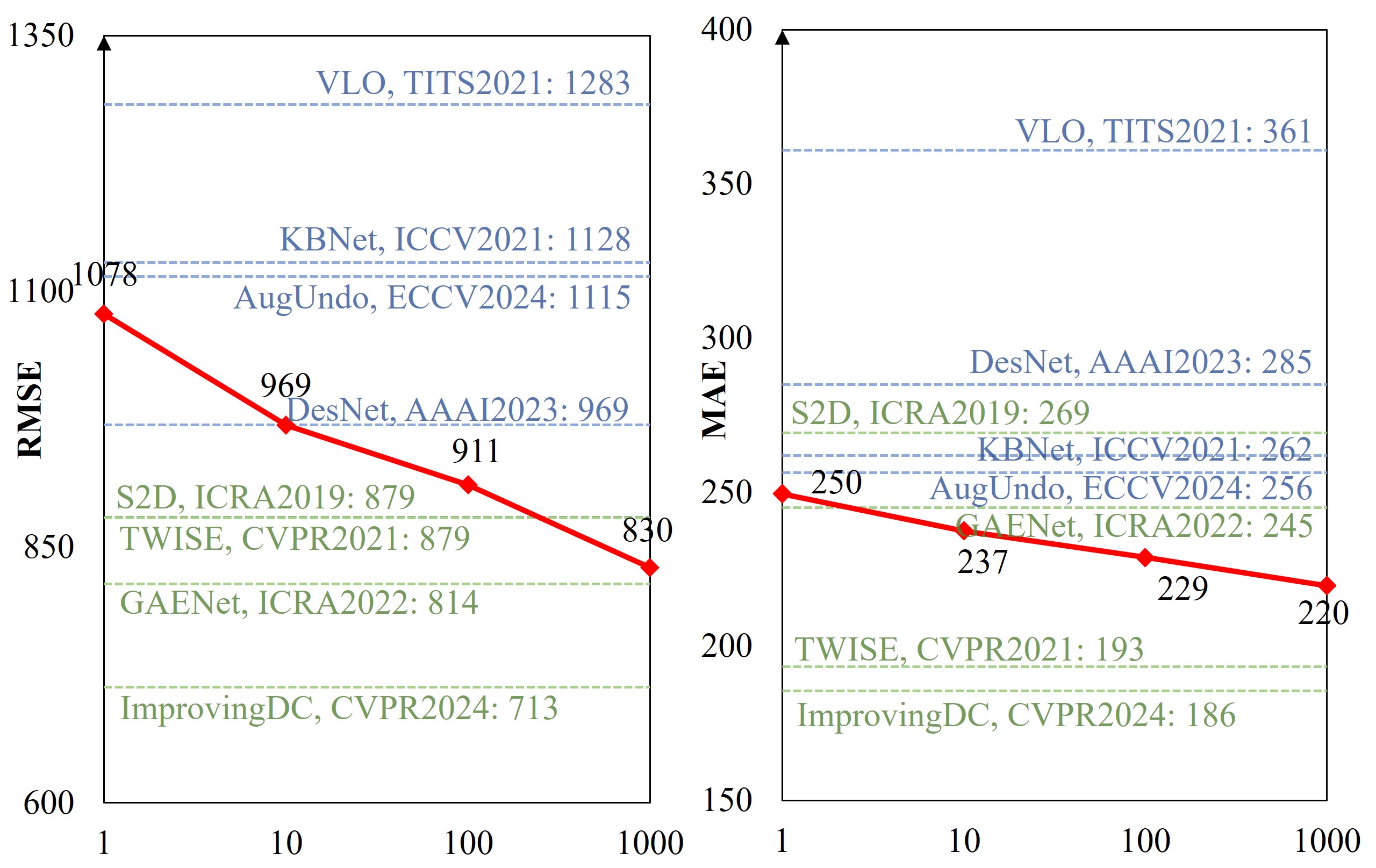}
   \caption{``\textbf{Few-shot vs full-shot}" on KITTI with 64 line LiDAR. The horizontal axis denotes that our models use 1, 10, 100, and 1000 training samples, while the baselines use 86K samples. The \textcolor{myblue}{self-supervised} baselines include VLO \cite{song2021self}, KBNet \cite{wong2021unsupervised}, AugUndo \cite{wu2024augundo}, and DesNet \cite{yan2023desnet}. The \textcolor{mygreen}{supervised} baselines include S2D \cite{ma2019self}, TWISE \cite{imran2021depth}, GAENet \cite{chen2022depth}, and ImprovingDC \cite{wang2024improving}.}
   \label{fig:fs-86k}
\end{figure}

\noindent\textbf{Baseline Details.} \textit{Zero-shot depth completion}: The baselines include several generalizable depth completion methods: G2-MonoDepth (G2MD) \cite{wang2023g2}, OMNI-DC \cite{zuo2024omni}, and SPNet \cite{wang2024scale}, as well as fully supervised methods: NLSPN \cite{park2020non}, CFormer \cite{zhang2023completionformer}, LRRU \cite{wang2023lrru}, and ImprovingDC \cite{wang2024improving}. To ensure zero-shot testing, the fully supervised methods are retrained on large-scale datasets provided by SPNet. \textit{Few-shot depth completion}: We retrained recent methods, LRRU \cite{wang2023lrru}, ImprovingDC \cite{wang2024improving}, and SparseDC \cite{long2024sparsedc} in the same few-shot setting for direct comparison. The official results of UniDC \cite{park2024simple} are listed for reference. To highlight our model, we further evaluate it, using less than 1000 samples, against full-shot baselines, trained with 86k samples. Their results are directly taken from the original papers.

\noindent\textbf{Metric Details.} We use standard evaluation metrics including root mean square error (RMSE), mean absolute error (MAE), root mean square error of the inverse depth (iRMSE), and mean absolute error of the inverse depth (iMAE). All results are reported in \textit{millimeters (mm)}.

\begin{table}[!t]
\centering
\resizebox{\linewidth}{!}{
\begin{tabular}{llcccc}
\toprule
Shot               & Methods  & \textit{RMSE} $\downarrow$ & \textit{MAE} $\downarrow$ & \textit{iRMSE} $\downarrow$ & \textit{iMAE} $\downarrow$ \\
\midrule
\multirow{4}{*}{\textit{1}}    & LRRU \cite{wang2023lrru}    & 5558 & 3708 & 259.59 & 23.60 \\
                      & ImprovingDC  \cite{wang2024improving} & 3322 & 1473 & 12.69  & 7.22 \\
                      & SparseDC \cite{long2024sparsedc} & 5950 & 3997 & 12598.67  & 250.22 \\
                      & \cellcolor{gray!15} \textbf{Ours}   & \cellcolor{gray!15} \textbf{1662} & \cellcolor{gray!15} \textbf{455} & \cellcolor{gray!15} \textbf{3.54}  & \cellcolor{gray!15} \textbf{1.42} \\
\midrule
\multirow{4}{*}{\textit{10}}   & LRRU  \cite{wang2023lrru} & 3206 & 1329 & 12.30  & 5.56 \\
                      & ImprovingDC   \cite{wang2024improving} & 3092 & 1350 & 13.06  & 6.20 \\
                      & SparseDC \cite{long2024sparsedc} & 3507 & 1659 & 50.56 & 9.61 \\
                      & \cellcolor{gray!15} \textbf{Ours}  & \cellcolor{gray!15} \textbf{1524} & \cellcolor{gray!15} \textbf{426} & \cellcolor{gray!15} \textbf{3.32}  & \cellcolor{gray!15} \textbf{1.37} \\
\midrule
\multirow{4}{*}{\textit{100}}  & LRRU \cite{wang2023lrru}   & 2646 & 1014 & 14.14  & 4.37 \\
                      & ImprovingDC  \cite{wang2024improving}  & 2642 & 1043 & 10.07  & 4.55 \\
                      & SparseDC \cite{long2024sparsedc} & 2235 & 798 & 19.30  & 3.47 \\
                      & \cellcolor{gray!15} \textbf{Ours}   & \cellcolor{gray!15} \textbf{1425}  & \cellcolor{gray!15} \textbf{339} & \cellcolor{gray!15} \textbf{3.07}  & \cellcolor{gray!15} \textbf{1.29} \\
\midrule
\multirow{4}{*}{\textit{1000}} & LRRU \cite{wang2023lrru}   & 2092 & 713 & 6.21  & 2.79 \\
                      & ImprovingDC  \cite{wang2024improving}   & 2259 & 843 & 7.82  & 3.67 \\
                      & SparseDC \cite{long2024sparsedc} & 1863 & 606 & 5.79  & 2.36 \\
                      & \cellcolor{gray!15} \textbf{Ours}  & \cellcolor{gray!15} \textbf{1297}  & \cellcolor{gray!15} \textbf{371} & \cellcolor{gray!15} \textbf{2.82}  & \cellcolor{gray!15} \textbf{1.22} \\
\bottomrule
\end{tabular}
}
\caption{\textbf{Few-shot depth completion} on KITTI with 64/32/16/8/4 lines LiDAR using 1, 10, 100, and 1000 training samples.}
\label{tab:fs-randline}
\end{table}

\begin{figure}[!t]
  \centering
   \includegraphics[width=\linewidth]{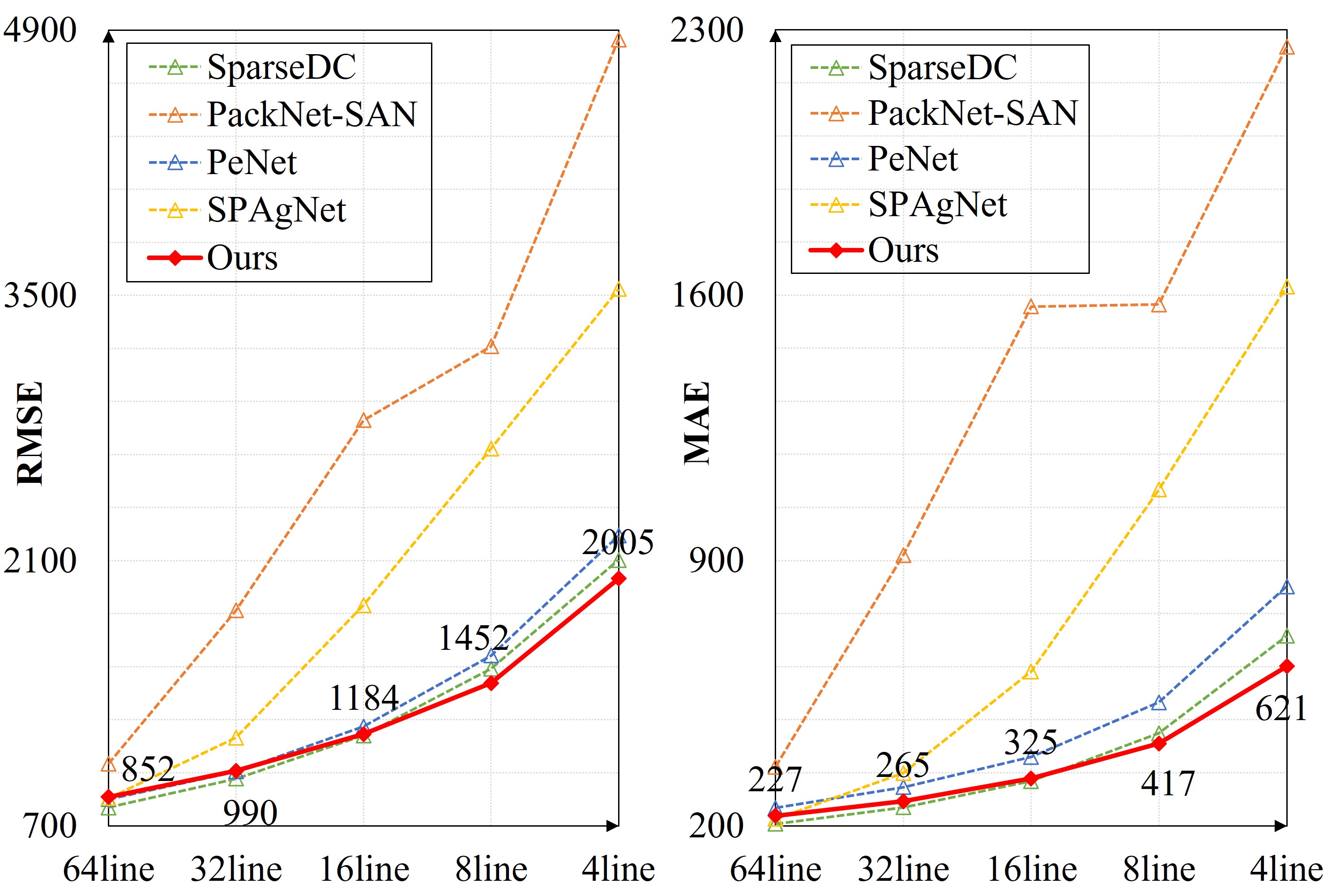}
   \caption{``\textbf{Few-shot vs full-shot}" on KITTI with 64/32/16/8/4 lines LiDAR. Our model is trained with 1000 samples, while the baselines use 86K samples including SparseDC \cite{long2024sparsedc}, PackNet-SAN \cite{guizilini2021sparse}, PeNet \cite{hu2021penet}, and SPAgNet \cite{conti2023sparsity}.}
   \label{fig:fs-randline}
\end{figure}

\begin{table*}[!t]
\centering
\scriptsize
\resizebox{\linewidth}{!}{
\begin{tabular}{lcc@{\hskip 5pt}c@{\hskip 5pt}cc@{\hskip 5pt}cc@{\hskip 5pt}c}
\toprule
\multirow{2}{*}{Framework} & Speed & \multicolumn{3}{c}{Basic Synthesis Method}       & \multicolumn{2}{c}{Extensions} & \multirow{2}{*}{RMSE $\downarrow$} & \multirow{2}{*}{MAE $\downarrow$} \\
\cmidrule(r){3-5} \cmidrule(r){6-7}
          &  (image/s)             & \textit{DepthAnything} \cite{yang2024depth} & \textit{P(Interpolation)} & \textit{Relocation} & \textit{DepthPro} \cite{bochkovskii2024depth}    & \textit{SA1B \textcolor{gray}{(subset)}} \cite{kirillov2023segment}    &                       &                      \\
\midrule
\multirow{8}{*}{\makecell[c]{SPNet \cite{wang2024scale} \\ \textcolor{gray}{\textit{(Tiny)}}}} & \multirow{8}{*}{126.6}  &               &               &            &             &                  & 2484                  & 990                  \\
             &            & $\checkmark$             & 0.0           &            &             &                  & 2463                  & 956                  \\
             &            & $\checkmark$             & 0.25        &            &             &                  & 2380                  & 912                  \\
             &            & $\checkmark$             & 0.5        &            &             &                  & 2344                  & 889                  \\
             &            & $\checkmark$             & 1.0           &            &             &                  & 2330                  & 889                  \\
             &            & $\checkmark$             & 1.0           & $\checkmark$          &             &                  & 2277                  & 857                  \\
             &            & $\checkmark$             & 1.0           & $\checkmark$          & $\checkmark$           &                  & 2241                  & 854                  \\
             &            & \cellcolor{gray!15}$\checkmark$             & \cellcolor{gray!15}1.0           & \cellcolor{gray!15}$\checkmark$          & \cellcolor{gray!15}$\checkmark$           & \cellcolor{gray!15}$\checkmark$                & \cellcolor{gray!15}\textbf{2143}                  & \cellcolor{gray!15}\textbf{792}                  \\
\midrule
\multirow{2}{*}{G2MD \cite{wang2023g2}}  &  \multirow{2}{*}{88.0}   &               &               &            &             &                  & 2669                  & 1101                 \\
             &            & \cellcolor{gray!15}$\checkmark$             & \cellcolor{gray!15}1.0           & \cellcolor{gray!15}$\checkmark$          & \cellcolor{gray!15}$\checkmark$           & \cellcolor{gray!15}$\checkmark$                & \cellcolor{gray!15}\textbf{2192}                  & \cellcolor{gray!15}\textbf{817}    \\
\bottomrule
\end{tabular}
}
\caption{\textbf{Ablation study} of our synthesis method based on two types of generalizable depth completion frameworks.}
\label{tab:zs-ab}
\end{table*}

\subsection{Zero-Shot Depth Completion}
\label{subsec:zero-shot}

In this section, we show model capability across zero-shot environments, which is unobserved during training.

\noindent\textbf{Uniformly Sampled Depths.}
We begin evaluation on sparse depth maps with uniformly sampled valid pixels at 10\%/1\%/0.1\% sparsity levels, across six unseen datasets. The average results are summarized in \cref{tab:zs-unf}, in which our model performs the best in most scenarios, with the lowest RMSE 1966 mm and MAE 731 mm on average. 

\noindent\textbf{Sensor Captured Depths.} 
Next, we evaluate sparse depth points captured by physical sensors on two unseen datasets: VOID \cite{wong2020unsupervised} dataset with 1500/500/150 sparsity levels, and KITTI \cite{geiger2012we} dataset with 64/16/4 lines LiDAR. \cref{tab:zs-lidar} shows that our model still outperforms all baseline models with the lowest RMSE 1147 mm and MAE 342 mm on average.

\noindent\textbf{Visual Results.} \cref{fig:teaser} has presented several visual results. Additional results can be found in supplementary material.

\subsection{Few-Shot Depth Completion}
\label{subsec:few-shot}

In this section, we fine-tune our model in the specific environment to compete for intro-domain learning methods, on KITTI dataset with 1, 10, 100, and 1000 samples. 

\noindent\textbf{64 line LiDAR Depths.} We train and test all models using 64 line LiDAR under the few-shot setting. \cref{tab:fs-64line} shows that our model significantly outperforms these baselines designed for intro-domain learning. We attribute this strong performance to powerful pre-trained weights by our synthesis method. These weights provide an excellent starting point to regularize this few-shot learning.

Additionally, we compare our few-shot model against baselines trained on the full 86K samples. As shown in \cref{fig:fs-86k}, our model achieves competitive results, outperforming all self-supervised baselines in MAE with just a single annotated sample. When using 1000 training samples, our model even outperforms supervised baselines such as S2D and TWISE in RMSE, as well as S2D and GAENet in MAE. 

\noindent\textbf{64/32/16/8/4 lines LiDAR Depths.} We also train all models on randomly sampled 64/32/16/8/4 lines LiDAR, and test them in each beam line. \cref{tab:fs-randline} shows the average results across these beam lines, and our models also outperform baseline methods. As shown in \cref{fig:fs-randline}, our few-shot model, trained with just 1000 samples, achieves competitive results of full-shot baselines. Notably, our model achieves the best RMSE and MAE on 8/4 lines LiDAR data.

\subsection{Ablation Study}
\label{subsec:ab}

In this section, we verify each component of our synthesis method. Due to computational constraints, we adopt the ``Tiny" model of SPNet using 25\% of the full data.

\noindent\textbf{Basic Synthesis Method.} Our synthesis method consists of three components to implement the basic approach outlined in \cref{eq:inter&reloc}: the incorporation of a depth foundation model (\textit{i.e.,} DepthAnything \cite{yang2024depth}), interpolation operation (\textit{i.e.,} P(Interpolation) = 1.0), and relocation operation. As shown in \cref{tab:zs-ab}, each of these steps progressively improves prediction quality by enhancing the geometry diversity.

\noindent\textbf{Extension Examples.} The basic synthesis method can be extended by integrating multiple depth foundation models and unlabeled data, as discussed in \cref{subsec:con&div}. For verification, we further incorporate an depth foundation model, DepthPro \cite{bochkovskii2024depth}, and 390K unlabeled images from SA1B \cite{kirillov2023segment}. The effectiveness of these extensions is shown in \cref{tab:zs-ab}. 
% While incorporating more depth foundation models and unlabeled data could theoretically further enhance performance, we do not explore this due to resource constraints.

\noindent\textbf{Interpolated vs Original Labels.} To further show the effectiveness of interpolated labels, we adjust the probability of using interpolated labels (\textit{i.e.,} P(Interpolation)) from 0.0 to 1.0. The remaining probability is used for randomly selecting original dense depth maps, including ground-truth depth maps and dense predictions from DepthAnything. As shown in \cref{tab:zs-ab}, when using fully interpolated depth labels (\textit{i.e.,} P(Interpolation) = 1.0), the model achieves the best performance due to maximized data diversity.

%% file: sec/5_conclusion.tex
\section{Conclusion}
\label{sec:conc}

In this paper, we propose a label-efficient method, PacGDC, to expand training data coverage with minimal annotation effort for generalizable depth completion. This is achieved by a data synthesis pipeline based on the 2D-to-3D projection ambiguities and consistencies. This pipeline includes multiple depth foundation models, interpolation and relocation strategies, and unlabeled data. Extensive experiments demonstrate that PacGDC achieves superior generalizable depth completion in both zero-shot and few-shot settings.

%% file: sec/6_acknowledgment.tex
\section*{Acknowledgment}

This work was supported in part by the National Natural Science Foundation of China (NSFC) under Grants 62088102 and 62373298, and by the Singapore Ministry of Education (MOE) Tier-2 Grant MOE-T2EP20123-0003.

%% file: sec/X_suppl.tex
\clearpage
\setcounter{page}{1}
\maketitlesupplementary

\section{More Implementation Details}
\label{sec:more_imp}

\noindent\textbf{Training Details.} \textit{Zero-shot Depth Completion:} The training data simply concentrate all available training datasets following \cite{wang2023g2, wang2024scale}, without any explicit balancing strategies as in \cite{zuo2024omni}. Due to resource constraints, the training resolution is set to 320$\times$320, though higher resolutions could further enhance performance, as observed in image analysis tasks \cite{liu2021swin, woo2023convnext}. 

Due to the challenges posed by our highly diverse data setting, we modify the 2$\times$2 convolutions in the up/downsampling layers of SPNet to 3$\times$3 convolutions, as odd-sized kernels generally provide better stability \cite{wu2019convolution}. This modification results in a slight increase in computational cost, as shown in \cref{tab:cost}. Notably, ``Ours-T" even outperforms ``SPNet-L" while requiring half the inference time and only 17\% of the parameters.

Additionally, we impose a constraint that the minimum number of sparse depth pixels during training is two. This allows us to simplify the absolute term in the G2-MonoDepth loss \cite{wang2023g2} to L1 loss. The updated loss function $\mathbb{L}$, which measures the discrepancy between predictions $\widetilde{d}$ and our pseudo depth labels $\hat{d}$, is expressed as follows:
\begin{equation}
\label{eq: g2loss}
\begin{aligned}
\mathbb{L}(\widetilde{d},\hat{d}) & =\frac{1}{\eta}{\sum_{i=1}^\eta}|T(\widetilde{d}_i)-T(\hat{d}_i)| + \frac{1}{\eta}{\sum_{i=1}^{\eta}}|\widetilde{d}_i-\hat{d}_i| \\
& + \frac{0.5}{\eta}{\sum_{r=0}^3}{\sum_{i=1}^\eta}|\nabla(\rho_r(T(\widetilde{d}_i)-T(\hat{d}_i)))|,
\end{aligned}
\end{equation}
where $T$ is the standardize operation with mean deviation in \cite{wang2023g2}. The function $\rho_r$ is the nearest neighbor interpolation at the $1/2^r$ resolution. $\nabla$ is the Sobel gradient in height and width directions. $\eta$ denotes the number of valid pixels in dense labels.

\textit{Few-shot Depth Completion:} In our few-shot experiments, we do not employ additional refinement strategies, such as SPN-like modules \cite{cheng2019learning, park2020non, wang2023lrru, tang2024bilateral} or depth enhancement methods \cite{wang2022depth,wang2023rgb}. This ensures that our model retains SPNet’s efficiency. The training resolution is set to a randomly cropped 256$\times$1216. The loss function is updated to the commonly used L1+L2 loss, following the standard practice in most intra-domain learning methods \cite{zhang2023completionformer, wang2023lrru, wang2024improving}.

\noindent\textbf{Testing Details.} The details of the test datasets are provided in \cref{tab:test_dataset}. For the uniform sampling experiment, test images are resized to a height of 320 pixels. In the sensor-captured experiment, the VOID and KITTI datasets follow standard protocols, with VOID maintaining its original resolution of 480$\times$640 and KITTI using a bottom center-cropped resolution of 256$\times$1216. The final results on KITTI are obtained by averaging predictions from both original and horizontally flipped inputs following implementations in \cite{wang2023lrru, wang2024improving}.

\begin{table}[!t]
\centering
\scriptsize
\resizebox{\linewidth}{!}{
\begin{tabular}{lc@{\hskip 5pt}ccc}
\toprule
Datasets & Indoor & Outdoor & Label & Size \\
\midrule
ETH3D \cite{schops2017multi} & \checkmark & \checkmark & Laser & 454 \\ 
Ibims \cite{koch2018evaluation} & \checkmark & & Laser & 100 \\
NYUv2 \cite{silberman2012indoor} & \checkmark & & RGB-D & 654 \\
DIODE \cite{vasiljevic2019diode} & \checkmark & \checkmark & Laser &  771 \\
Sintel \cite{butler2012naturalistic} & \checkmark & \checkmark & Synthetic & 1064 \\
KITTI \cite{geiger2012we} &  & \checkmark & Stereo & 1000 \\
VOID \cite{wong2020unsupervised} & \checkmark & \checkmark & RGB-D & 800 \\
\bottomrule
\end{tabular}
}
\caption{The details of test datasets.}
\label{tab:test_dataset}
\end{table}

\begin{table}[!t]
\centering
\resizebox{\linewidth}{!}{
\begin{tabular}{cccccc}
\toprule
\multirow{2}{*}{Method} & Speed$\uparrow$ & Param.$\downarrow$ & Memo.$\downarrow$ & RMSE$\downarrow$ & MAE$\downarrow$  \\
                        & (Image/s)  & (M)       & (MB)   & (mm) & (mm) \\
\midrule
SPNet-T              & \textbf{126.6}      & \textbf{35.0}       & 330    & 2342 & 857  \\
Ours-T             & 121.8      & 39.7       & \textbf{242}    & \textcolor{gray}{\textbf{2143}} & \textcolor{gray}{\textbf{792}}  \\
\midrule
SPNet-L             & \textbf{60.2}       & \textbf{235.5}     & \textbf{1176}   & 2271 & 791  \\
Ours-L              & 58.7       & 254.4    & 1246   & \textbf{1966} & \textbf{731} \\
\bottomrule
\end{tabular}
}
\caption{The inference costs under ``Tiny" (T) and ``Large" (L) configurations, including speed, parameters, and memory usage. Notably, the results of ``Ours-T" are copied from the ablation study only using 25\% training data (in \textcolor{gray}{gray}).}
\label{tab:cost}
\end{table}

\section{More Quantitative Results}
\label{sec:more_quan}
\noindent\textbf{Zero-shot Depth Completion on DDAD Dataset.} We further evaluate PacGDC on the DDAD \cite{guizilini20203d} dataset, comparing to more generalizable and supervised baselines, following the standard protocol of VPP4DC \cite{bartolomei2024revisiting}. The baseline results are directly taken from relevant papers. As shown in \cref{tab:DDAD}, the results further validate the effectiveness of PacGDC for zero-shot generalization.

\begin{table}[h]
\centering
\resizebox{\linewidth}{!}{
\begin{tabular}{lcclcc}
\toprule
Method & \textit{RMSE} $\downarrow$ & \textit{MAE} $\downarrow$ & Method & \textit{RMSE} $\downarrow$ & \textit{MAE} $\downarrow$ \\
\midrule
BP-Net \cite{tang2024bilateral}  & 8903 & 2712 & Marigold-DC \cite{viola2024marigold} & 6449 & 2364\\
VPP4DC \cite{bartolomei2024revisiting}   & 10247 & 2290  & DMD$^3$C \cite{liang2025distilling}  & 6609 & 1842  \\
OGNI-DC \cite{zuo2024ogni}   & 6876 & 1867 & \cellcolor{gray!20} \textbf{Ours} & \cellcolor{gray!20} \textbf{5918} & \cellcolor{gray!20} \textbf{1140} \\
\bottomrule
\end{tabular}
}
\caption{\textbf{Zero-shot depth completion} on DDAD dataset under VPP4DC protocol.}
\label{tab:DDAD}
\end{table}

\noindent\textbf{Few-shot Comparison with Other Baselines.} We supplement \cref{tab:fs-64line} with additional few-shot baselines. The baseline results on KITTI validation set are directly taken from their original papers. As shown in \cref{tab:few-shot_more}, the results further demonstrate the superiority of our model in few-shot depth completion.

\begin{table}[h]
\centering
\resizebox{\linewidth}{!}{
\begin{tabular}{lccccc}
\toprule
Method & Shot & \textit{RMSE} $\downarrow$ & \textit{MAE} $\downarrow$ & \textit{iRMSE} $\downarrow$ & \textit{iMAE} $\downarrow$ \\
\midrule
DepthPrompt \cite{park2024depth}  & 100 & 1798 & 602 & - & - \\
\rowcolor{gray!20} \textbf{Ours} & 100 & \textbf{911} & \textbf{229} & \textbf{2.54} & \textbf{0.96} \\
\midrule
DDPMDC \cite{ran2023few}  & 11000 & 966 & 291 & 3.63 & 1.48 \\
\rowcolor{gray!20} \textbf{Ours} & 1000 & \textbf{830} & \textbf{220} & \textbf{2.28} & \textbf{0.91} \\
\bottomrule
\end{tabular}
}
\caption{\textbf{Few-shot depth completion} on KITTI with 64 line LiDAR, supplementing \cref{tab:fs-64line}.}
\label{tab:few-shot_more}
\end{table}

\noindent\textbf{In-Domain Evaluation on the KITTI Dataset.} We further conduct standard in-domain evaluation by fine-tuning the pre-trained zero-shot PacGDC model on the entire KITTI training set (\textit{i.e.,} 86K samples). As presented in \cref{tab:in-domain}, despite adopting a plain backbone without specialized components such as spatial propagation networks (SPNs), PacGDC delivers competitive performance on the KITTI validation set, comparable to recent state-of-the-art methods. Moreover, we submit the results of the fully fine-tuned model to the official KITTI test set leaderboard.

\begin{table}[h]
\centering
\resizebox{\linewidth}{!}{
\begin{tabular}{lccccc}
\toprule
Method & Plain & \textit{RMSE} $\downarrow$ & \textit{MAE} $\downarrow$ & \textit{iRMSE} $\downarrow$ & \textit{iMAE} $\downarrow$ \\
\midrule
BEV@DC \cite{zhou2023bev}  & & 720 & \textbf{187} & \textbf{1.88} & \textbf{0.80} \\
TPVD \cite{yan2024tri}  & & \textbf{719} & 187 & - & - \\
\midrule
BEV@DC \cite{zhou2023bev}  & $\checkmark$ & 762 & 198 & 2.06 & 0.86 \\
TPVD \cite{yan2024tri}  & $\checkmark$ & 764 & \textbf{198} & - & - \\
UniDC \cite{park2024simple} & $\checkmark$ & 824 & 209 & - & - \\
\rowcolor{gray!20} \textbf{Ours} & $\checkmark$ & \textbf{759} & 203 & \textbf{2.06} & \textbf{0.85} \\
\bottomrule
\end{tabular}
}
\caption{\textbf{In-domain evaluation} on KITTI validation set.}
\label{tab:in-domain}
\end{table}

\section{More Ablation Study}
\label{sec:more_abl}

\noindent\textbf{Different Depth Foundation Models.} We evaluate our approach with four different depth foundation models: DepthAnything (DA) \cite{yang2024depth}, DepthPro \cite{bochkovskii2024depth}, DepthAnythingV2 (DAV2) \cite{yang2024depthv2}, and DistillAnyDepth (DistillAD) \cite{he2025distill}. As shown in \cref{tab:fundation}, PacGDC consistently yields performance improvements over the baseline (without PacGDC), further validating the generality and effectiveness of our method.

It is worth noting that this experiment was newly introduced in response to reviewer feedback. Accordingly, our "Large" model continues to use DA and DepthPro, as reported in \cref{tab:zs-ab}, rather than the combination of DA, DepthPro, and DAV2 used in \cref{tab:fundation}.

\begin{table}[h]
\centering
\resizebox{\linewidth}{!}{
\begin{tabular}{cccccc}
\toprule
DA \cite{yang2024depth} & DepthPro \cite{bochkovskii2024depth} & DAV2 \cite{yang2024depthv2} & DistillAD \cite{he2025distill} & \textit{RMSE} $\downarrow$ & \textit{MAE} $\downarrow$ \\
\hline
& & & &  2484 & 990 \\
\rowcolor{gray!10} $\checkmark$ & & & &   2277 & 857 \\
\rowcolor{gray!10} $\checkmark$ & $\checkmark$ & & &   2241 & 854  \\
\rowcolor{gray!10} & $\checkmark$ & $\checkmark$ &  & 2243  & 852  \\
\rowcolor{gray!10} & & $\checkmark$ & $\checkmark$ & 2276  &  859 \\
\rowcolor{gray!10} $\checkmark$ & $\checkmark$ & $\checkmark$ & & \textbf{2232} & \textbf{848}   \\
\rowcolor{gray!10} $\checkmark$ & $\checkmark$ & $\checkmark$ & $\checkmark$ & 2279  & 874  \\
\bottomrule
\end{tabular}
}
\caption{\textbf{Ablation study} on different depth foundation models. Results with our data synthesis pipeline are shaded in \textcolor{gray}{gray}.}

\label{tab:fundation}
\end{table}

\section{More Visual Results}
\label{sec:more_qual}

\begin{figure*}[!t]
    \centering
    \includegraphics[width=0.95\linewidth]{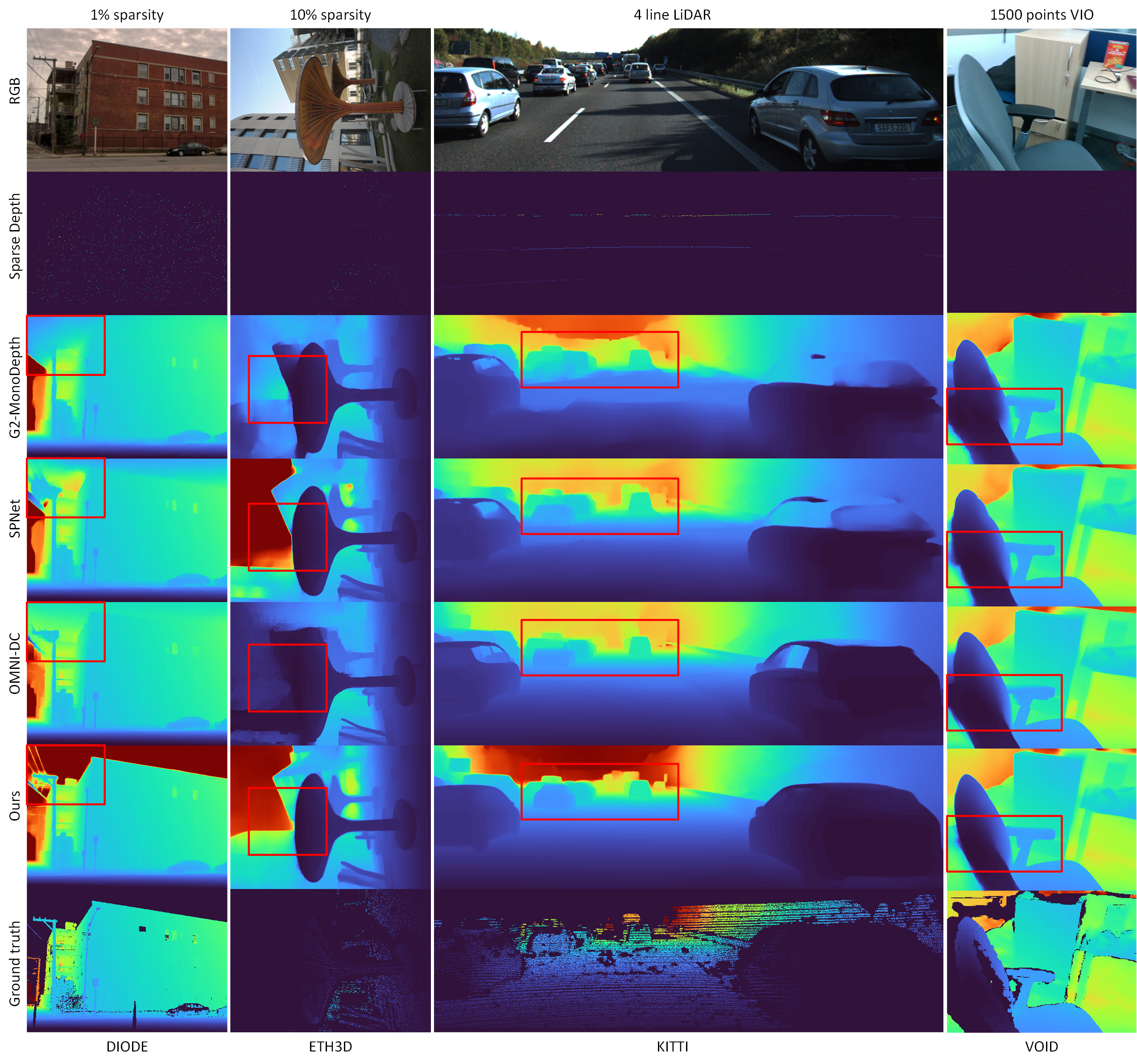}
    \caption{\textbf{Zero-shot depth completion} on unseen scenarios with different scene semantics/scales and depth sparsity/patterns. \label{fig: visual_zs}}
\end{figure*}

\noindent\textbf{Zero-Shot Depth Completion.} We further provide visual examples of zero-shot scenarios in \cref{fig: visual_zs}, covering a range of datasets and sparsity levels: DIODE with 1\% sparsity, ETH3D with 10\% sparsity, KITTI with 4-line LiDAR, and VOID with 1500 feature points derived from a VIO system. Across these scenarios, characterized by diverse scene semantics, varying scales, and different forms of depth sparsity, PacGDC consistently achieves higher accuracy in predicting metric depth maps compared to existing baselines.

\begin{figure*}[!t]
    \centering
    \includegraphics[width=0.95\linewidth]{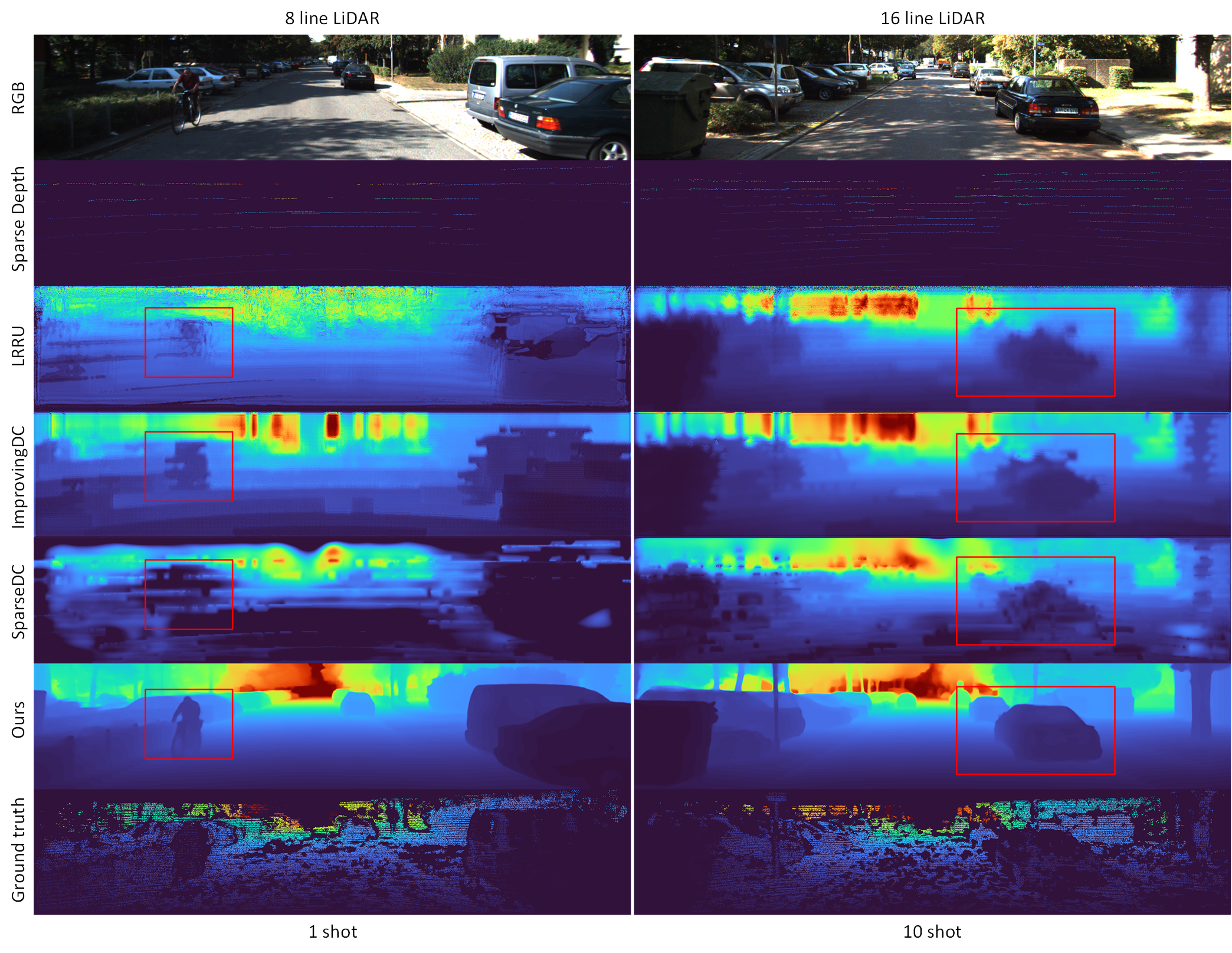}
    \caption{\textbf{Few-shot depth completion} on KITTI with 8- and 16-lines LiDAR, using models trained with 1 and 10 samples, respectively. \label{fig: visual_fs_110}}
\end{figure*}

\begin{figure*}[!t]
    \centering
    \includegraphics[width=0.95\linewidth]{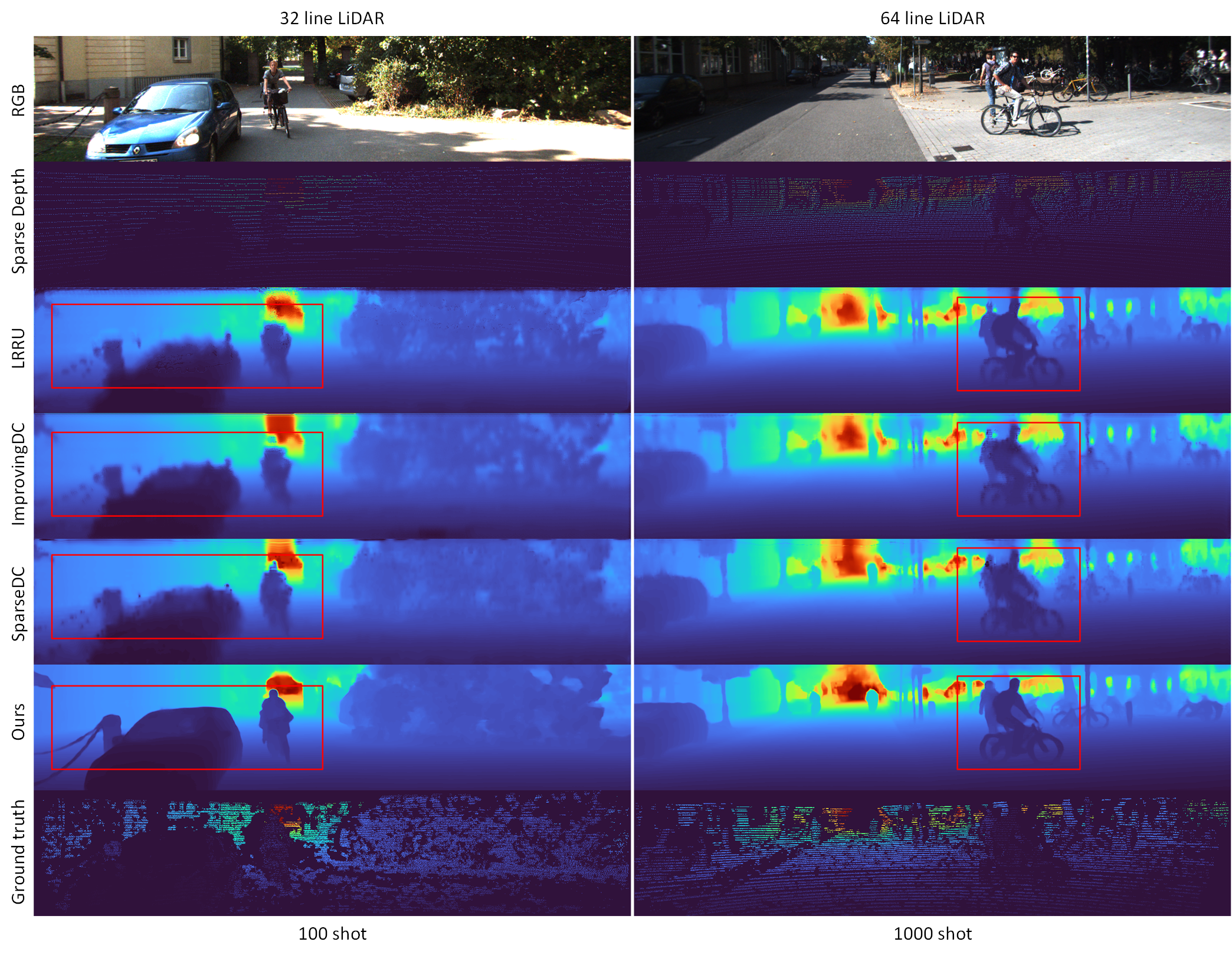}
    \caption{\textbf{Few-shot depth completion} on KITTI with 32- and 64-lines LiDAR, using models trained with 100 and 1000 samples, respectively. \label{fig: visual_fs_1001000}}
\end{figure*}

\noindent\textbf{Few-Shot Depth Completion.} Visual results for few-shot scenarios are presented in \cref{fig: visual_fs_110,fig: visual_fs_1001000}, using models trained with 1, 10, 100, and 1000 samples. To provide a comprehensive analysis, we also separately showcase results for 8-, 16-, 32-, and 64-line LiDAR inputs under the same few-shot training settings. Leveraging the strong pre-trained weights from our synthesis pipeline, our model demonstrates significant qualitative improvements over in-domain learning baselines across all levels of supervision.

